\pdfoutput=1

\documentclass[11pt]{article}

\usepackage{acl}

\usepackage{times}
\usepackage{latexsym}

\usepackage[T1]{fontenc}

\usepackage[utf8]{inputenc}

\usepackage{microtype}

\usepackage{subcaption}

\usepackage{url}
\usepackage{graphicx}
\usepackage{graphics}
\usepackage{amsmath}
\usepackage{amsthm}
\usepackage{algorithm}
\usepackage{algorithmic}
\usepackage{microtype}
\usepackage{booktabs} 
\usepackage{makecell}
\usepackage{arydshln}   
\usepackage{multirow}
\usepackage{bbm}
\usepackage{balance}
\usepackage{amssymb}
\usepackage{textcomp}
\usepackage{color}
\usepackage{xcolor}
\usepackage{enumitem}
\usepackage{CJKutf8}

\usepackage[export]{adjustbox}

\usepackage{hyperref}

\definecolor{zred}{RGB}{196, 38, 11}
\definecolor{zblue}{RGB}{41, 52, 190}
\definecolor{zgreen}{RGB}{18, 141, 21}

\definecolor{zptu}{RGB}{18, 141, 21}

\newcommand{\oaxe}{\textsc{OaXE}~}

\title{On the Information Redundancy in Non-Autoregressive Translation}
\author{Zhihao Wang$^{1,3}$, Longyue Wang$^{2}$, Jinsong Su$^{1,3}$, Junfeng Yao$^3$, Zhaopeng Tu$^2$ \\
        $^1$School of Informatics, Xiamen University, China \\
        $^2$Tencent AI Lab, China \\        
        $^3$Key Laboratory of Digital Protection and Intelligent Processing of Intangible Cultural Heritage \\of Fujian and Taiwan (Xiamen University), Ministry of Culture and Tourism, China\\
        \texttt{zhwang@stu.xmu.edu.cn}~~~\texttt{\{vinnylywang,zptu\}@tencent.com}\\ \texttt{\{jssu,yao0010\}@xmu.edu.cn}}

\begin{document}

\maketitle

\begin{abstract}
Token repetition is a typical form of multi-modal problem in fully non-autoregressive translation (NAT).
In this work, we revisit the multi-modal problem in recently proposed NAT models. Our study reveals that these advanced models have introduced other types of information redundancy errors, which cannot be measured by the conventional metric -- the continuous repetition ratio.
By manually annotating the NAT outputs, we identify two types of information redundancy errors that correspond well to {\em lexical} and {\em reordering} multi-modality problems.
Since human annotation is time-consuming and labor-intensive, we propose automatic metrics to evaluate the two types of redundant errors. Our metrics allow future studies to evaluate new methods and gain a more comprehensive understanding of their effectiveness.

\end{abstract}

\section{Introduction}
\label{intro}

Fully non-autoregressive translation (NAT) has received increasing attention for its efficient decoding by predicting every target token in parallel~\cite{NAT}.
However, such advantage comes at the cost of sacrificing translation quality due to the {\em multi-modality} problem:
there exist many possible translations of the same sentence, while vanilla NAT models may consider them at the same time due to the independent predictions, 
which leads to multi-modal outputs in the form of token repetitions~\cite{NAT}.
Accordingly,~\newcite{maskp} propose an automatic metric to measure the multi-modality problem with the {\bf continuous repetition ratio} in the generated output, which is widely-used in the following works.


Recent works have incorporated approaches to ameliorate the effect of multi-modality with a proxy of target distribution~\cite{huang:2022:learning}. For example, a thread of research enhances the decoder input by masking or deleting partial target sentence~\cite{maskp,glat,zheng2023towards}. Another thread of work modifies the target labels to adaptively adjust to the model output~\cite{imputer,axe,oaxe,ngram-oaxe,dat}.
These works have reported to greatly alleviate the multi-modality problem in terms of significantly reducing the continuous repetition ratio in the generated output.

\begin{table}[t!]
\centering
\scalebox{0.85}{
\begin{tabular}{ccc}
\toprule
\bf Type  & \bf Example \\
\midrule
Continuous Repetition     & I {\em ate} {\bf ate} pizza tonight.\\
Discontinuous Repetition  & {\em Tonight} I ate pizza {\bf tonight}.\\
Continuous Synonym        & I {\em ate} {\bf had} pizza tonight.\\
Discontinuous Synonym     & I {\em ate} pizza {\bf had} tonight.\\
\bottomrule
\end{tabular}}
\caption{Illustration of information redundancy cases. Redundant tokens are highlighted in {\bf bold}. Conventional token repetition metric only considers the continuous repetition, while ignores the other types of redundancy.}
\label{tab:redundancy-cases}
\end{table}

\begin{table*}
\centering
\begin{tabular}{l rr rrrr}
\toprule
\multirow{2}{*}{\bf Model}   & \multirow{2}{*}{\bf BLEU}  &  \bf Cont.  &    \multicolumn{4}{c}{\bf Other Redundancy Errors}\\
\cmidrule(lr){4-7}
    &  &   \bf  Rep.  & \em Total  &  \em Cont. Syn.   & \em Disc.  Rep.  &  \em Disc. Syn.\\
\cmidrule(lr){1-3}\cmidrule(lr){4-7}
\multicolumn{7}{l}{\bf Raw Data}\\
CMLM{\small~\cite{maskp}}      & 15.0 & 47.36\% & 3.66\% & 0.55\% & 3.08\% & 0.03\%\\
\hdashline
~~~+ \textsc{OaXE}~\cite{oaxe}     & 36.9 & 6.20\% & 3.40\% & 1.53\% & 1.56\% & 0.31\%\\
GLAT~\cite{glat}               & 35.9 & 6.78\% & 2.70\% & 1.24\% & 1.17\% & 0.29\%\\
~~~+ \textsc{OaXE}~\cite{oaxe}     & 36.7 & 3.89\% & 2.91\% & 1.11\% & 1.45\% & 0.35\%\\  
DAT~\cite{dat}                 & 40.8 & 0.00\% & 7.14\% & 0.28\% & 5.75\% & 1.11\%\\
\cmidrule(lr){1-3}\cmidrule(lr){4-7}
\multicolumn{7}{l}{\bf Distilled Data}\\
CMLM{\small~\cite{maskp}}      & 27.5 & 22.13\% & 3.29\% & 0.41\% & 2.32\% & 0.56\%\\
\hdashline
~~~+ \textsc{OaXE}~\cite{oaxe}      & 39.6 & 4.30\% & 2.35\% & 0.71\% & 1.05\% & 0.59\% \\
GLAT~\cite{glat}               & 40.5 & 2.59\% & 2.05\% & 0.57\% & 1.23\% & 0.25\%\\
~~~+ \textsc{OaXE}~\cite{oaxe}      & 41.1 & 2.55\% & 1.95\% & 0.50\% & 1.20\% & 0.25\%\\
DAT~\cite{dat}                 & 43.4 & 0.00\% & 4.95\% & 0.43\% & 3.18\% & 1.34\%\\
\bottomrule
\end{tabular}
\caption{Human evaluation on 100 sentences sampled from WMT20 English$\rightarrow$Chinese testset. 
Continuous repetition (``Cont. Rep.'') is the conventional metric to measure the redundancy information of the NAT output. ``Total'' denotes the total ratio of the other redundancy errors.}
\label{tab:human-evaluation_bpe}
\end{table*}

However, we find that these advanced models have introduced other types of information redundancy errors, which cannot be measured by the conventional repetition metric. By carefully examining the outputs of advanced NAT models, we summarize three other types of information redundancy errors, which correspond well to different multi-modality problems (as listed in Table~\ref{tab:redundancy-cases}):
\begin{itemize}[leftmargin=10pt]
    \item {\em Continuous {\bf Synonym}}: While the conventional metric only considers repetition of the same token (e.g., ``ate'' and ``ate''), this category includes synonyms (different words with the same meaning, e.g., ``ate'' and ``had''), which corresponds to {\em lexical multi-modality} in translation. 
    \item {\em {\bf Discontinuous} Repetition}: Another common source of multi-modality is word reordering~\cite{oaxe}, which leads to discontinuous repetition. For example, ``tonight'' may be at the beginning of a sentence in one modality, and at the end in another modality.
    \item {\bf Discontinuous Synonym}: This category is a combination of the above two types of errors, which is a mixture of lexical and reordering multi-modalities.
\end{itemize}

We manually annotate the four types of redundant errors in 100 sentences for five representative NAT models. We find that the multi-modality problem in DAT~\cite{cmlmc} is mainly in the form of discontinuous redundancy, while that in \textsc{OaXE}~\cite{oaxe} and GLAT~\cite{glat} is mainly in the form of continuous redundancy (\S~\ref{sec:revisit}). 
Since manually annotating redundant errors is time-consuming and labor-intensive, we propose automatic metrics to evaluate redundant errors, which correlates well with human annotation (\S\ref{sec:automatic-metric}).
The automatic metric enables us to build NAT benchmarks for information redundancy, which we hope can facilitate future research along this direction (\S\ref{sec:benchmarks}).

\begin{table*}[t]
\small
    \centering
    \begin{tabular}{l m{13.8cm}}
    \toprule
    \bf Source & Vis@@ as in the desert kingdom , endowed with rich be@@ dou@@ in heritage and archaeological sites , are currently restricted to exp@@ at workers , their depen@@ dents and Muslim pilgri@@ ms travelling to holy sites in Mec@@ ca and Med@@ ina . \\
    \hdashline
    \bf Refer. & \begin{CJK}{UTF8}{gbsn}在 这个 拥有 众多 贝@@ 多 因 遗迹 和 考古 遗址 的 沙漠 王国 ， 目前 签证 仅限于 外籍 雇员 、 他们 的 家属 和 前往 麦加 和 麦@@ 地@@ 那 圣地 的 穆斯林 朝圣者 。\end{CJK}\\
    \midrule
    \bf CMLM & \begin{CJK}{UTF8}{gbsn}沙漠 王国 {\color{zred}的 的 的} {\color{zblue}丰富 丰富 丰富} {\color{zred}的 的} {\color{zgreen}和 和} 考古 {\color{zred}， ，} {\color{zblue}目前 目前} {\color{zgreen}被 被} 工人 {\color{zred}的} 、 {\color{zred}的} {\color{zblue}和 和} {\color{zgreen}穆斯林} 前往 {\color{zred}麦加 麦加 麦加 麦加 }地@@ 那 的 {\color{zgreen}穆斯林} 圣地 。\end{CJK}\\
    \hdashline
    \bf ~~+\oaxe & \begin{CJK}{UTF8}{gbsn}沙漠 王国 中 的 签证 拥有 {\color{zred}丰富 丰富} 贝都 因人 遗产 和 考古 遗址 ， 目前 只 限于 被 驱逐 工人 、 他们 的 家属 和 前往 麦加 和 麦@@ 地@@ 那 圣地 的 穆斯林 朝圣者 。\end{CJK}\\
    \hdashline
    \bf GLAT & \begin{CJK}{UTF8}{gbsn}在 沙漠 王国 ， 拥有 了 丰富 的 {\color{zred}贝都 都} 遗产 和 遗址 的 {\color{zblue}签证 签证} 目前 只 限于 外籍 工人 、 他们 的 {\color{zgreen}家属 家属} 和 前往 麦加 和 麦@@ 地@@ 那 圣地 的 穆斯林 朝圣者 。\end{CJK}\\
    \hdashline
    \bf ~~+\oaxe & \begin{CJK}{UTF8}{gbsn}在 沙漠 王国 {\color{zred}， ，} 拥有 丰富 {\color{zred}贝都 都} 遗产 和 考古 遗址 的 签证 目前 只 限于 外籍 工人 、 他们 的 家属 和 {\color{zblue}前往 前往} 麦加 和 麦@@ 地@@ 那 的 穆斯林 朝圣者 。\end{CJK}\\
    \hdashline
    \bf DAT & \begin{CJK}{UTF8}{gbsn}沙漠 王国 的 签证 拥有 丰富 的 {\color{zred} 贝都 都} 因 遗产 和 考古 遗址 ， 目前 只 限于 {\color{zblue}外籍 工人 、 他们 的 家属 和 穆斯林 朝圣者} 前往 麦加 和 麦@@ 地@@ 那 圣地 的 {\color{zblue}外籍 工人 、 其 家属 和 穆斯林 朝圣者} 。\end{CJK}\\
    \bottomrule
    \end{tabular}
    \caption{Examples of English$\rightarrow$Chinese translation on raw data. Redundancy tokens are highlighted in {color}.}
    \label{tab:case}
\end{table*}

\section{Revisiting Redundancy in NAT}
\label{sec:revisit}

In this section, we revisit the information redundancy problem in fully NAT models.

\paragraph{Models}
We use CMLM~\cite{maskp} as our baseline, which enhances decoder input by sampling the number of masked tokens from a fixed uniform distribution.
We re-implement several representative NAT models on top of CMLM:
\begin{itemize}[leftmargin=10pt]
    \item \textsc{OaXE}~\cite{oaxe} removes the penalty of word order errors with a novel order-agnostic cross entropy loss, which enables NAT models to handle word reordering.
    \item GLAT~\cite{glat} adaptively masks the target sentence to adjust the training difficulties.
    \item DAT~\cite{dat} further introduces a directed acyclic graph to simultaneously capture multiple translations on top of GLAT.
\end{itemize}
We follow~\newcite{scaling4nat} to train CMLM, GLAT and DAT models for 300K steps with a batch size of 480K. We train \textsc{OaXE} models by fine-tuning from trained CMLM and GLAT models for 10 epochs. 
All NAT models are re-implemented on top of the Fairseq~\cite{fairseq} framework.
We evaluate the translation performance on an ensemble of 5 best checkpoints (ranked by validation BLEU~\cite{bleu}). 
More details of the experimental setup  are shown in Appendix~\ref{sec:experimental-setup}.

\paragraph{Human Annotation} We sample 100 sentences from the WMT20 English$\rightarrow$Chinese testset, and manually annotate the four types of redundant errors in the outputs generated by NAT models. The annotation guideline is listed in Appendix~\ref{sec:annotation-guideline}.

\paragraph{Results}
Table~\ref{tab:human-evaluation_bpe} lists the results of human evaluation on WMT20 English$\rightarrow$Chinese (En$\rightarrow$Zh, 21.8M). Clearly, all methods significantly improve translation performance (e.g. 20+ BLEU on raw data and 10+ BLEU on distilled data) and greatly reduce the continuous repetition ratio, which is consistent with the results reported in the original papers. DAT reduces continuous repetition ratio to 0\% with a transition probability matrix, which would assign low probability to the edge of two repetitive tokens (i.e. continuous repetition).

However, by examining the outputs closely, we found that the multi-modality problem in DAT is mainly in the form of discontinuous redundancy (e.g. repetition and synonym). One possible reason is that the transition probability matrix in DAT is a first-order HMM and cannot assign probability to discontinuous tokens. In contrast, the redundancy problems in \textsc{OaXE} and \textsc{GLAT} are mainly in the continuous form. 
We attribute this phenomenon to that both approaches ease the training difficulty while lack an explicit strategy (e.g. transition matrix in DAT) to model the dependency between two neighbouring tokens.
Knowledge distillation reduces the redundancy errors, while the trend still holds in distilled data.
Table~\ref{tab:case} shows some examples of redundant outputs on raw data.\footnote{More translation examples can be found in Appendix~\ref{sec:examples}.}

\section{Automatic Evaluation}

One of the great challenges in human evaluation lies in that manually annotating redundant errors is time-consuming and labor-intensive. 
In response to this problem, we propose automatic metrics to evaluate redundant errors, allowing future work to evaluate new methods to obtain a more complete picture of a method's effectiveness.

\subsection{Automatic Metric}
\label{sec:automatic-metric}

For simplicity, we combine repetition and synonym for continuous and discontinuous redundancies.

\paragraph{Redundancy Checking} Two tokens are redundant if they are repeated or a pair of synonyms:
\begin{align*}
    R(y_i, y_j) = 
    \begin{cases}
        1, &\text{if $y_i=y_j$ or $S(y_i, y_j)=1$}\\
        0, &\text{otherwise}
    \end{cases}
\end{align*}
where $S(y_i, y_j)=1$ denotes $y_i$ and $y_j$ are a pair of synonyms. We leverage word embedding to construct synonyms. Specifically, we use the multilingual word embedding in pretrained mBART~\cite{mbart} model to broad applicability of our metric across diverse languages. Two tokens are a pair of synonyms if the similarity of their word embeddings is greater than threshold value.

\begin{table}[t!]
\renewcommand\thetable{4}
\centering
\scalebox{0.92}{
\begin{tabular}{c ccc ccc}
\toprule
\bf Train  & \multicolumn{3}{c}{\bf Continuous}  & \multicolumn{3}{c}{\bf Discontinuous} \\
\cmidrule(lr){2-4}\cmidrule(lr){5-7}
\bf Data    &   P   &   R   &   F1  &   P   &   R   &   F1\\
\midrule
Raw         & 99.1 & 96.9 & 98.0 & 75.1 & 86.5 & 80.4 \\
Distilled   & 98.8 & 96.2 & 97.5 & 64.2 & 91.4 & 75.4 \\
\bottomrule
\end{tabular}
}
\caption{Evaluation of automatic metric on human-annotated NAT outputs for WMT20 En$\rightarrow$Zh translation.}
\label{tab:metric-evaluation}
\end{table}

\begin{table*}[t!]
\renewcommand\thetable{5}
\centering
\setlength{\tabcolsep}{5pt}
\begin{tabular}{l rrr rrr rr}
\toprule
\multirow{2}{*}{\bf Model}   & \multicolumn{2}{c}{\bf W20 En$\rightarrow$Zh}   & \multicolumn{2}{c}{\bf W20 Zh$\rightarrow$En}   & \multicolumn{2}{c}{\bf W20 En$\rightarrow$De}   & \multicolumn{2}{c}{\bf W20 De$\rightarrow$En}\\
\cmidrule(lr){2-3}\cmidrule(lr){4-5}\cmidrule(lr){6-7}\cmidrule(lr){8-9}
    & \em BLEU  & \em Redu. &  \em BLEU  & \em Redu. &  \em BLEU  & \em Redu. &  \em BLEU  & \em Redu.\\
\midrule
\multicolumn{9}{l}{\bf Raw Data}\\
CMLM{\small~\cite{maskp}}      & 13.7 & 53.5\% &  7.6 & 55.0\% &  9.3 & 42.4\% & 10.3 & 50.8\% \\
\hdashline
~~~+ \textsc{OaXE}~\cite{oaxe} & 33.9 & 11.6\% & 19.3 &  8.8\% & 22.9 &  6.1\% & 32.2 &  6.7\% \\
GLAT~\cite{glat}               & 33.3 & 12.2\% & 20.7 & 10.5\% & 24.0 &  5.9\% & 36.1 &  5.6\% \\
~~~+ \textsc{OaXE}~\cite{oaxe} & 35.3 &  8.7\% & 21.1 &  7.3\% & 25.9 &  5.2\% & 35.6 &  5.3\% \\  
DAT~\cite{dat}                 & 39.4 &  8.1\% & 25.8 &  5.5\% & 29.4 &  3.6\% & 39.6 &  3.8\% \\
\midrule
\multicolumn{9}{l}{\bf Distilled Data}\\
CMLM{\small~\cite{maskp}}      & 27.1 & 27.6\% & 16.4 & 30.9\% & 20.7 & 19.3\% & 24.2 & 23.3\% \\
\hdashline
~~~+ \textsc{OaXE}~\cite{oaxe} & 37.5 &  9.1\% & 24.5 &  8.4\% & 28.4 &  4.5\% & 38.2 &  4.7\% \\
GLAT~\cite{glat}               & 38.6 &  7.0\% & 26.4 &  6.2\% & 30.6 &  3.7\% & 40.0 &  3.8\% \\
~~~+ \textsc{OaXE}~\cite{oaxe} & 39.3 &  6.9\% & 26.4 &  5.8\% & 30.7 &  3.5\% & 40.3 &  3.7\% \\
DAT~\cite{dat}                 & 42.0 &  5.8\% & 28.5 &  5.1\% & 32.7 &  3.1\% & 41.5 &  3.3\% \\
\bottomrule
\end{tabular}
\caption{Information redundancy benchmarks for fully NAT models on large-scale WMT20 datasets. Detailed results of information redundancy can be found in Table~\ref{tab:detailed-benchmark} in Appendix.}
\label{tab:benchmark}
\end{table*}

\paragraph{Continuous Redundancy}
The continuous redundancy is straightforward to calculate by checking whether two neighbouring tokens are redundant:
\begin{align*}
    \textsc{Cr}(y_i) = 
    \begin{cases}
        1, &\text{if $i>1$ and $R(y_i, y_{i-1})=1$}\\
        0, &\text{otherwise}
    \end{cases}
\end{align*}

\paragraph{Discontinuous Redundancy}
The discontinuous redundancy cannot be computed by simply checking whether each token is redundant with previous tokens. While there are hardly two continuous repeated tokens in natural sentences, discontinuous tokens are grammatically correct in certain cases. Take the source sentences in Table~\ref{tab:case} for example, there are several discontinuous redundant function words such as ``and'', ``in'', and ``,''. In addition, the content word ``sites'' is used twice in different clauses. Accordingly, two exceptional cases are not considered as discontinuous redundancy:
\begin{CJK}{UTF8}{gbsn}
\begin{itemize}[leftmargin=10pt]
    \item {\em Stopwords} $S$: We consider the top-3 frequent tokens in training data (e.g. ``，'', ``。'', and ``的'' in Chinese) as stopwords, which are often used several times in a sentence.
    \item {\em Tokens with Multiple Occurrences} $\Tilde{M}$: A redundant token would be exempt if it or its synonym occur multiple times in the reference or source sentence, and the exempt quota is no more than the number of repeated occurrence. To this end, we maintain a dynamic list $\Tilde{M}$, which removes the redundant token when it hits the list.
\end{itemize}
\end{CJK}

For the tokens that are not in the exceptional cases (i.e. $y_i \notin S \cup \Tilde{M}$), we calculate the discontinuous redundancy by:
\begin{align*}
    \textsc{Dr}(y_i) = 
    \begin{cases}
        1, &\text{if $\exists {j<i}, R(y_i, y_j)=1$}\\
        0, &\text{otherwise}
    \end{cases}
\end{align*}

The ratios of continuous and discontinuous redundancy errors are calculated by normalizing the \textsc{Cr} and \textsc{Dr} values by the number of all tokens:
\begin{align*}
    \textsc{Crr}(Y) = {\sum_{y_i \in Y}\textsc{Cr}(y_i)}/{(|Y|-1)} \\
    \textsc{Drr}(Y) = {\sum_{y_i \in Y}\textsc{Dr}(y_i)}/{(|Y|-1)}
\end{align*}
Table~\ref{tab:metric-evaluation} lists the accuracy of the automatic metrics on the human annotated NAT outputs, which demonstrates the reliability of the proposed metrics. The accuracy of the discontinuous metric is relatively lower, since not all discontinuous redundancy tokens are mistaken.
Table~\ref{tab:human-automatic} in Appendix lists the detailed results of automatic metrics, which share similar trend with human evaluation across models.

\subsection{Information Redundancy Benchmarks}
\label{sec:benchmarks}

With the automatic metrics in hand, we are able to evaluate the information redundancy for more language pairs. Table~\ref{tab:benchmark} lists the results on the full testsets of large-scale WMT20 En$\leftrightarrow$Zh (21.8M) and En$\leftrightarrow$De (45.1M) translation tasks. 
Clearly, all advanced NAT models consistently improve translation performance over the CMLM baseline by reducing the information redundancy ratio. However, there is still an improvement space for all advanced models, indicating that current NAT models still suffer from the multi-modality problem.

\section{Conclusion}

In this paper, we revisit the multi-modality problem in fully NAT, and identify two new types of information redundancy errors that correspond to lexical and reordering multi-modality problems. 
We propose automatic metrics to evaluate these redundant errors and establish benchmarks for information redundancy, allowing future work to evaluate new methods more comprehensively. 

\section*{Limitations}

The limitations of this work are:
\begin{itemize}[leftmargin=10pt]
    \item The conclusions in this paper are drawn from several representative NAT models on large-scale datasets, which are not necessarily well suited for other NAT models and datasets (e.g. small-scale WMT16 En-Ro).
    \item The synonym list is derived from mBART embedding, which may not be accurate. In the future, we will develop new strategies to build high-quality synonym lists across languages.
\end{itemize}

\section*{Ethics Statement}
We take ethical considerations very seriously, and strictly adhere to the ACL Ethics Policy.
This paper focuses on the information redundancy problem in the outputs of NAT models, which can be seen as a linguistic problem. 
Both the datasets and models used in this paper publicly available and have been widely adopted by researches of non-autoregressive translation. We ensure that the findings and conclusions of this paper are reported accurately and objectively.

\bibliography{ref}
\bibliographystyle{acl_natbib}

\clearpage
\appendix

\section{Human Annotation Guideline}
\label{sec:annotation-guideline}

In general, the process of manual labeling totally spends two professional translators (native target-language speakers) for two months, which costs US \$5,000 dollars. We annotated the widely-used WMT testset at BPE-token level. The annotation guidelines are as follows:
\begin{CJK}{UTF8}{gbsn}
\begin{enumerate} [leftmargin=*,topsep=0.1em,itemsep=0.1em,parsep=0.1em]
  \item Each instance contains three parts: {\em source input} ({\bf src}), {\em system output} ({\bf sys}) and {\em reference} ({\bf ref}). The annotators first detect all possible redundant tokens by comparing {\bf sys} (detailed in Section~\ref{sec:revisit})  and {\bf ref}, and further select candidates that fit the Redundancy Definition (defined in Section~\ref{intro}) by reviewing {\bf src}.

  
  \item As we define four types of information redundancy, when processing a instance one annotator should scan four times, each one of which only targets one redundancy type.

  
  \item The annotator check neighbouring tokens for ``Continuous'' types but should consider the whole sentence for ``Discontinuous'' ones. The ``Repetition'' tokens can be easily detected according to surface form while the ``Synonym'' requires semantic analysis. We therefore suggest to evaluate the candidates in terms of {\em fluency} (from a native speaker's point of view) and {\em adequacy} (from a translator's point of view), which are usually strong indicators of synonym redundancy. 
  
  
  \item Each redundancy case can be represented into a tuple $(a,b)_{n}^{m}$, where $a$ and $b$ are position IDs of mutually redundant tokens. Besides, $n$ is the redundancy type (e.g. Discontinuous Synonym) and $m$ denotes the NAT system (e.g. DAT trained on KD dataset). This format facilitates quantification such as distribution statistics and automatic evaluation. 
  
  
  
\end{enumerate}
\end{CJK}
 
\noindent{We sampled 100 instances from the WMT2020 English$\rightarrow$Chinese testset for each NAT system. As we investigated five NAT models (i.e. CMLM, CMLM+\textsc{OaXE}, GLAT, GLAT+\textsc{OaXE}, DAT) and two data settings (i.e. Raw and KD), each individual totally annotates 1000 instances. We followed \citet{mitani2017summary} to measure the inter-annotator agreement by calculating the average pairwise cohen's kappa score. This reaches 0.86\% kappa scores, demonstrating that the annotators work efficiently and consistently under this guideline.} This is potentially useful for annotating more datasets in other languages (e.g. German$\rightarrow$English).

\section{Experimental Setup}
\label{sec:experimental-setup}



\begin{table*}[t!]
\centering
\begin{tabular}{l rr rrr}
\toprule
\multirow{2}{*}{\bf Model} & \multirow{2}{*}{\bf Size} & \multirow{2}{*}{\bf GPU Hours} & \multicolumn{3}{c}{\bf Hyper-Parameters} \\
 & & & Learning Rate & Dropout & Label Smoothing \\
\midrule
CMLM                & 87M &  320h & 7e-4 & 0.0   & 0.1 \\
~~~+ \textsc{OaXE}~ & 87M &   24h & 5e-6 & 0.0   & 0.1 \\
GLAT~               & 89M &  840h & 5e-4 & 0.0   & 0.1 \\
~~~+ \textsc{OaXE}~ & 89M &   24h & 5e-6 & 0.0   & 0.1 \\
DAT~                & 90M & 2240h & 5e-4 & 0.1   & 0.0   \\
\bottomrule
\end{tabular}
\caption{The number of parameters, training budget (in GPU hours for NVIDIA A100), and training hyper-parameters of different NAT models. More detailed training hyper-parameters can be found in the training scripts of the different NAT models, which will be released.}
\label{tab:parameters}
\end{table*}

\subsection{Training}

In order to compare different NAT models more fairly, we unify some important training hyper-parameters in the training process. We adopt Transformer-Base configuration for all NAT models: both encoder and decoder contain 6 layers with 8 attention heads, the hidden dimension is 512, and the feedforward layer dimension is 2,048. We train all NAT models with a big batch size of 480K. We train CMLM, GLAT and DAT models for 300K steps, while train \textsc{OaXE} models by fine-tuning from trained CMLM and GLAT models for 10 epochs. 

Other important hyper-parameters are listed in Table~\ref{tab:parameters}.
More detailed training hyper-parameters can be found in the training scripts of the different NAT models, which will be released.

\subsection{Decoding}
All NAT models used in this paper are fully NAT models. Hence, the decoding iteration for all NAT model is 1. We use a length beam size of $b = 1$ for all NAT models, apart from CMLM ($b = 5$). Specially, we adopt greedy decoding algorithm for DAT. More detailed of decoding can be found in the decoding scripts of the different NAT models, which will be released.

\subsection{Automatic Metric}

\paragraph{mBART} The synonym list is derived from mBART embedding, which may be wrong. For example, the embeddings of top frequent token have high similarity with all other embeddings. Hence, we also consider the top-3 frequent chinese tokens in mBART dict as stop words. For other non-Chinese languages, we set the top-10 frequent non-Chinese tokens as stop words.

\paragraph{Stopwords} Unlike Chinese, English and German have more shared tokens for different words. Hence, we also set top-10 requent tokens in training data of other language (eg. English and German).

\section{Translation Examples for Information Redundancy}
\label{sec:examples}

In this section, we include more examples of translation from WMT20 English$\rightarrow$Chinese testset. Table~\ref{tab:case1} and Table~\ref{tab:case2} show the source sentence, reference and translations of different NAT models. Instances of redundancy tokens are highlighted in red color.

\section{Detailed Results of Information Redundancy}
\label{sec:benchmarks}

In this section, we provide more detailed results of information redundancy. The detailed results of human and automatic evaluation are listed in Table~\ref{tab:human-automatic}. It compares the results of human and automatic evaluations on 100 sentences from WMT20 En$\rightarrow$Zh testset. Besides, we also provide the detailed information redundancy benchmarks, which are shown in Table~\ref{tab:detailed-benchmark}.

\begin{table*}[t]
\small
    \centering
    \begin{tabular}{l m{13.8cm}}
    \toprule
    \bf Source & The government has compul@@ sor@@ ily retired 15 more tax officers in the fourth tran@@ che of its crack@@ down on err@@ ant officials accused of corruption and other mal@@ practices .\\
    \hdashline
    \bf Refer. & \begin{CJK}{UTF8}{gbsn}政府 在 第四@@ 轮 打击 涉嫌 腐败 和 其他 渎职 行为 的 失职 官员 行动 中 ， 又 强制 退休 了 15 名 税务 官员 。\end{CJK}\\
    \midrule
    \bf CMLM & \begin{CJK}{UTF8}{gbsn}在 {\color{zred} 对 对 腐败 腐败 腐败 腐败 腐败 腐败 的 的 官员 官员 官员 官员 官员} 的 {\color{zred} ， ， ， 政府 政府 15 15 15 官员 官员 官员 官员} 。\end{CJK}\\
    \hdashline
    \bf ~~+\oaxe & \begin{CJK}{UTF8}{gbsn}政府 在 第四阶段 打击 对 {\color{zred}被控 指控} 腐败 和 其他 不良@@ 行为 的 偏离 使命 的 官员 的 行动 中 政府 已 {\color{zred}强制 强制} 了 15 名 税务 官员 退休 。\end{CJK}\\
    \hdashline
    \bf GLAT & \begin{CJK}{UTF8}{gbsn}政府 已 强制 退休 了 15 名 税务 官员 ， 对 {\color{zred}被 被} 指控 腐败 和 其他 渎职 {\color{zred}的 的} {\color{zred}官员} 的 第四@@ 的 予以 {\color{zred}官员} 。\end{CJK}\\
    \hdashline
    \bf ~~+\oaxe & \begin{CJK}{UTF8}{gbsn}政府 已经 在 第四 阶段 强制 退休 15 名 税务 官员 ， 打击 被 指控 为 腐败 和 其他 渎职 行为 {\color{zred}的 官员 的 的 官员} 。\end{CJK}\\
    \hdashline
    \bf DAT & \begin{CJK}{UTF8}{gbsn}{\color{zred} 政府} 已 {\color{zred} 强制 又 退休 了 15 名 税务 官员 ，} 在 对 被控 腐败 和 其他 不当 行为 的 渎职 官员 的 镇压 {\color{zred} 行动} 的 第四 批 {\color{zred} 行动} 中 ， {\color{zred} 政府 又 强制 退休 了 15 名 税务 官员 。}\end{CJK}\\
    \bottomrule

    \toprule
    \bf Source & Antonio Brown has indicated he 's not retiring from the NF@@ L , only a few days after announcing he was done with the league in a Twitter rant .\\
    \hdashline
    \bf Refer. & \begin{CJK}{UTF8}{gbsn}安东尼奥 · 布朗 表示 他 不会 从 美国 国家 橄榄球 联盟 退役 ， 就 在 几天 前 ， 他 在 推@@ 特@@ 上 嚷@@ 嚷 不休 地 宣布 他 已经 离开 国家 橄榄球 联盟 。\end{CJK}\\
    \midrule
    \bf CMLM & \begin{CJK}{UTF8}{gbsn}安东尼奥 · 布朗 表示 {\color{zred}他 他 他 联盟 联盟 联盟 联盟 ， ， 几天 几天 ， ， 他 他 他 他 他 联盟 联盟 联盟 联盟 联盟} 。\end{CJK}\\
    \hdashline
    \bf ~~+\oaxe & \begin{CJK}{UTF8}{gbsn}安东尼奥 · 布朗 已经 表示 他 {\color{zred}不会 不会} 从 N@@ F@@ 联盟 退休 就 在 几天 后 他 在 Twitter {\color{zred}上 中} 宣布 他 在 了 联盟 的 结束 。\end{CJK}\\
    \hdashline
    \bf GLAT & \begin{CJK}{UTF8}{gbsn}安东尼奥 布朗 表示 他 不会 从 N@@ F@@ L 退休 ， 仅仅 几天 前 {\color{zred}在 在 上 上 中} 宣布 他 已经 {\color{zred}联盟 联盟 联盟 联盟} 。\end{CJK}\\
    \hdashline
    \bf ~~+\oaxe & \begin{CJK}{UTF8}{gbsn}安东尼奥 布朗 表示 他 没有 从 N@@ F@@ L 退役 {\color{zred}， ， 在 几天 在 在} 特@@ {\color{zred}上 中} 宣布 他 {\color{zred}在} 联盟 {\color{zred}的 的 几天} 。\end{CJK}\\
    \hdashline
    \bf DAT & \begin{CJK}{UTF8}{gbsn}安东尼奥 · 布朗 表示 {\color{zred}他 不会 从 N@@ F@@ L 退休} ， 仅仅 在 {\color{zred}几天 后 ，} 他 {\color{zred}宣布 他} 在 Twitter 上 的 咆哮 中 {\color{zred}宣布 他} 与 联盟 结束 {\color{zred}后 几天 后} ， {\color{zred}他} 并 {\color{zred}不会 从 N@@ F@@ L 退休} 。\end{CJK}\\
    \bottomrule

    \toprule
    \bf Source & " He knew how to manipulate the media . He knew exactly how to get the front page , " F@@ id@@ des , who was Jackson 's body@@ guard for 10 years , said . " 90 per cent of the time it worked , by putting a mask on his face , or stic@@ ky tape on his hands - or tape on his nose was a favourite one . He would say he wanted his life to be the greatest mystery on Earth .\\
    \hdashline
    \bf Refer. & \begin{CJK}{UTF8}{gbsn}" 他 知道 如何 操纵 媒体 。 他 完全 知道 如何 登上 头@@ 条@@ 新闻 ， " 做 了 杰克逊 10 年 保镖 的 菲 德斯 说道 。 " 脸上 戴上 口@@ 罩 ， 或者 用 胶@@ 带 粘 在 手上 ， 或者 用 胶@@ 带 粘 在 鼻子 上 （ 这个 他 最 喜欢 做 ） ， 在 90\% 的 情况 下 ， 这种 方法 很 奏效 。 他会 说 ， 他 希望 自己 的 一生 成为 全球 最大 的 谜 。 "\end{CJK}\\
    \midrule
    \bf CMLM & \begin{CJK}{UTF8}{gbsn}{\color{zred}杰克逊 杰克逊 的 的 的 的 的 的 的 的 的 " " " 的 的 的 的 的 的 ， ， ， ， ， 的 的 的 ， ， ， ， ， ， ， ， ， ， ， ， ， ， ， ， ， ， 的 的 的 的 的 的 的 的 的 的 的 的 的 他 他 他 他 他 他 的 的 的} 成为 世界 上 最 {\color{zred}的 的 的} 。 "\end{CJK}\\
    \hdashline
    \bf ~~+\oaxe & \begin{CJK}{UTF8}{gbsn}曾 担任 杰克逊 10 年 的 保镖 的 菲@@ {\color{zred}德斯 德斯} 说 ： " 他 知道 如何 {\color{zred}媒体 媒体} ， {\color{zred}他 他} 知道 如何 {\color{zred}头@@ 头@@ 。 。} " {\color{zred}90\% 90\%} 的 时间 都 是 在 给 他 {\color{zred}上 上} 面具 ， 或者 胶@@ 带 粘 在 {\color{zred}手上 手上} ， -- 或者 用 带 在 的 鼻子 上 最 喜欢 {\color{zred}的 的} ， 他 说 {\color{zred}希望 希望} 自己 的 生命 成为 地球 上 最 伟大 的 秘密 。 "\end{CJK}\\
    \hdashline
    \bf GLAT & \begin{CJK}{UTF8}{gbsn}" 他 知道 操纵 媒体 ， 他 知道 如何 登上 {\color{zred}头@@ 头@@} 版 " 做 杰克逊 了 10 年 {\color{zred}的 的} 菲@@ 德 {\color{zred}说 说 " "} 90\% 的 工作 的 时间 ， 在 他 脸上 了 一个 面具 ， 或者 手上 有 粘@@ 胶@@ 带 ， 在 鼻子 {\color{zred}上 上} 是 {\color{zred}最 最} 的 {\color{zred}， ，} 他 说 希望 他 能 成为 地球 最 伟大 的 神秘 。\end{CJK}\\
    \hdashline
    \bf ~~+\oaxe & \begin{CJK}{UTF8}{gbsn}" 他 知道 如何 操纵 媒体 ， 他 知道 {\color{zred}如何 如何} 登上 头@@ 版 ， " 作为 杰克逊 的 保镖 了 {\color{zred}十年 年} 的 菲@@ 德 {\color{zred}说 说} ， " {\color{zred}90\% 90\%} 的 工作 时间 ， 在 {\color{zred}脸上} 戴 {\color{zred}上} 一个 面具 ， {\color{zred}手上 手上} 的 胶@@ 带 ， 或者 在 鼻子 上 的 胶@@ 带 是 最 喜欢 的 。 他 说 他 希望 他 生命 成为 地球 上 最 伟大 的 神秘 。 "\end{CJK}\\
    \hdashline
    \bf DAT & \begin{CJK}{UTF8}{gbsn}{\color{zred}" 他 知道} 如何 {\color{zred}操纵 媒体 ， 他} 非常 {\color{zred}知道 如何 获得 头@@ 版 ，} " 杰克逊 的 保镖 10 年 的 菲@@ 德 斯 {\color{zred}说} ： {\color{zred}" 他 懂得 操纵 媒体 ， 他 知道 如何 获得 头@@ 版 ， 他 说} ， " 90\% 的 时间 里 ， {\color{zred}他} 都 通过 在 {\color{zred}他} 脸上 戴 面具 ， 或者 在 {\color{zred}手上} 贴 {\color{zred}上} 一张 {\color{zred}胶@@ 胶} 带子 ， 或者 在 他 鼻子 上 的 带子 ， 是 他 最 喜欢 的 方式 ， {\color{zred}他} 将 {\color{zred}说} ， 他 希望 他 的 一生 成为 地球 上 最 神秘 的 人 。 "\end{CJK}\\
    \bottomrule
    \end{tabular}
    \caption{Examples of En$\rightarrow$Zh translation on raw data. Redundancy errors are highlighted in {\color{zred} red color}.}
    \label{tab:case1}
\end{table*}

\begin{table*}[t]
\small
    \centering
    \begin{tabular}{l m{13.8cm}}
    \toprule
    \bf Source & They were addressed to her son , who has autism and lives in a private care facility , she said . But instead of her son 's name inside when you opened them , the letters said Dear Maine 's Department of Health and Human Services -- in C@@ incin@@ nat@@ i , she told local media .\\
    \hdashline
    \bf Refer. & \begin{CJK}{UTF8}{gbsn}这些 信@@ 是 邮寄 给 她 患有 自闭症 住 在 私人 护理 院 的 儿子 的 ， 她 解释 说 。 但 拆@@ 开 信件 ， 信@@ 里 未 提及 她 儿子 的 名字 ， 而是 说 " 尊敬 的 缅因州 卫生 与 公众 服务部 -- 辛辛@@ 那@@ 提 " ， 她 告诉 当地 媒体 。\end{CJK}\\
    \midrule
    \bf CMLM & \begin{CJK}{UTF8}{gbsn}她 说 ， 信 {\color{zred}的 的 的 的 的 的 的 ， ， ， ， ， ， ， ， ， ， ， ， ， ， ， ， 的 的 的 的 的 的 的 的 的 的 的 的 的 ， ， ， ， 的 的 的 的 ， ， 的 的} 她 {\color{zred}的 的} 姓名 。\end{CJK}\\
    \hdashline
    \bf ~~+\oaxe & \begin{CJK}{UTF8}{gbsn}她 说 ， 信 {\color{zred}是 是} 给 她 儿子 ， 他 患有 孤独@@ 症 ， 住 在 私人 护理 机构 里 {\color{zred}， ， ，} 但 位于 辛辛@@ 那@@ 提 的 梅@@ 缅因州 卫生 与 公众 服务部 的 {\color{zred}告诉 告诉} 当地 媒体 说 ， {\color{zred}在 在 时 时 她 的 的 名字} 而 不是 {\color{zred}她} 儿子 {\color{zred}的 名字} 。\end{CJK}\\
    \hdashline
    \bf GLAT & \begin{CJK}{UTF8}{gbsn}她 说 ， 信 是 给 她 患有 自闭症 并 生活 在 私人 护理 机构 的 儿子 {\color{zred}写 写} 的 {\color{zred}， ， 不是 不是} 打开 信 {\color{zred}时 时} 她 儿子 的 名字 ， 而是 信中 亲爱 的 缅因州 {\color{zred}卫生 和 卫生 与} 服务部 的 -- 她 向 当地 媒体 说 。\end{CJK}\\
    \hdashline
    \bf ~~+\oaxe & \begin{CJK}{UTF8}{gbsn}她 说 ， 这些 是 针对 她 患有 自闭症 ， 住 在 私人 护理 机构 儿子 的 ， 但 不是 你 打开 信 {\color{zred}的} 时候 {\color{zred}的} 她 儿子 的 名字 ， 她 {\color{zred}说 说} 亲爱 的 缅因州 {\color{zred}卫生 卫生} 与 服务部 ， 辛辛@@ 那@@ {\color{zred}提 提} 对 当地 媒体 说 。\end{CJK}\\
    \hdashline
    \bf DAT & \begin{CJK}{UTF8}{gbsn}她 说 ， 这些 信 是 写给 {\color{zred}她 的 儿子} ， {\color{zred}她} 患有 {\color{zred}自闭症} 并 {\color{zred}住 在 私人 护理} 机构 {\color{zred}的 儿子} ， 但 当 你 打开 {\color{zred}信 时} ， {\color{zred}她} 的 {\color{zred}信} 并 {\color{zred}没有} 以 {\color{zred}她 儿子 的 名字 ， 而是 说} ， {\color{zred}辛辛@@ 那@@ 提 亲爱 的} 缅因州 {\color{zred}卫生 和} 公众 {\color{zred}服务部 说} ， 她 对 当地 媒体 {\color{zred}说} ， 这些 {\color{zred}封信} 中 是 {\color{zred}告诉 亲爱 的 辛辛@@ 那@@ 提 的 卫生 和 服务部 的 时} ， {\color{zred}而 在 她 的} ， {\color{zred}她 自闭症} ， {\color{zred}住 在 私人 护理 里} ， {\color{zred}而 不是 在 里面 没有 她 儿子 的 名字} 。\end{CJK}\\
    \bottomrule
    
    \toprule
    \bf Source & HSBC was forced to invest hundreds of millions of dollars in financial crime compliance as part of a \$ 1.9 billion settlement in 2012 with U.S. authorities over the bank 's failure to prevent money laundering by drug cartels though its Mexican unit .\\
    \hdashline
    \bf Refer. & \begin{CJK}{UTF8}{gbsn}2012 年 ， 汇@@ 丰@@ 银行 由于 未能 阻止 毒@@ 贩 通过 其 墨西哥 子公司 进行 洗钱 活动 ， 与 美国 当局 达成 一项 19 亿美元 的 和解 协议 。 其中 被迫 在 金融 犯罪 合规 方面 投资 的 数@@ 亿美元 就是 该 协议 的 一部分 。\end{CJK}\\
    \midrule
    \bf CMLM & \begin{CJK}{UTF8}{gbsn}2012 {\color{zred}年 年} 与 {\color{zred}年} 19 {亿美元 亿美元 亿美元} ， {\color{zred}洗钱 洗钱 洗钱 洗钱 洗钱 的 的 的 洗钱 洗钱 洗钱 洗钱 ， ，} 汇@@ {\color{zred}被迫 被迫 被迫 集团 集团 集团 洗钱} 了 的 一部分 。\end{CJK}\\
    \hdashline
    \bf ~~+\oaxe & \begin{CJK}{UTF8}{gbsn}汇@@ 丰 被迫 投入 数@@ 亿美元 用于 遵守 金融 犯罪 合规 ， 作为 2012 年 与 美国 当局 19 {\color{zred}亿美元 亿美元} 因 {\color{zred}墨西哥 墨西哥 洗钱} 问题 未能 防止 {\color{zred}墨西哥} 毒@@ 集团 {\color{zred}洗钱 洗钱} 解决方案 的 一部分 。\end{CJK}\\
    \hdashline
    \bf GLAT & \begin{CJK}{UTF8}{gbsn}汇@@ 丰@@ 银行 被迫 投入 数@@ 亿美元 用于 金融 犯罪 {\color{zred}合规 合规} ， 作为 2012 {\color{zred}年 年} 与 美国 当局 {\color{zred}和解 和解} 的 一部分 ， 因为 该 银行 未能 通过 其 墨西哥 部门 来 防止 毒品 卡特尔 的 洗钱 。\end{CJK}\\
    \hdashline
    \bf ~~+\oaxe & \begin{CJK}{UTF8}{gbsn}汇@@ 丰@@ 银行 被迫 投入 数@@ 亿美元 的 {\color{zred}金融 金融} 的 遵守 性 ， 作为 2012 年 {\color{zred}与} 19 亿美元 {\color{zred}与} 美国 和解 {\color{zred}一部分} ， 因 银行 未能 {\color{zred}防止} 其 墨西哥 部门 的 {\color{zred}防止} 毒@@ 集团 洗钱 的 {\color{zred}一部分} 。\end{CJK}\\
    \hdashline
    \bf DAT & \begin{CJK}{UTF8}{gbsn}汇@@ 丰@@ 银行 被迫 投资 数@@ 亿美元 用于 遵守 金融 犯罪 ， 作为 2012 年 {\color{zred}与 美国 当局 达成 19 亿美元 和解 的 一部分 ，} 原因 是 汇@@ 丰@@ 银行 未能 通过 其 墨西哥 子公司 防止 贩@@ 毒@@ 集团 洗钱 ， {\color{zred}与 美国 当局 达成 的 19 亿美元 和解 的 一部分 。}\end{CJK}\\
    \bottomrule
    \end{tabular}
    \caption{Examples of En$\rightarrow$Zh translation on raw data. Redundancy errors are highlighted in {\color{zred} red color}.}
    \label{tab:case2}
\end{table*}

\begin{table*}
\centering
\begin{tabular}{l rr rr}
\toprule
\multirow{2}{*}{\bf Model}   &    \multicolumn{2}{c}{\bf Continuous Redundancy}   &    \multicolumn{2}{c}{\bf Discontinuous Redundancy}\\
\cmidrule(lr){2-3}\cmidrule(lr){4-5}
    &  \em Human   &    \em Automatic   &  \em Human   &    \em Automatic\\
\midrule
\multicolumn{5}{l}{\bf Raw Data}\\
CMLM{~\cite{maskp}}         & 47.91\% & 47.81\% & 3.11\% & 2.11\%\\
\hdashline
~~~+ \textsc{OaXE}~\cite{oaxe}    &  7.73\% & 7.07\% & 1.87\% & 3.58\%\\
GLAT~\cite{glat}                  &  8.02\% & 7.42\% & 1.46\% & 3.14\%\\
~~~+ \textsc{OaXE}~\cite{oaxe}    &  4.99\% & 4.61\% & 1.80\% & 3.63\%\\  
DAT~\cite{dat}                    &  0.28\% & 0.40\% & 6.87\% & 7.39\%\\
\midrule
\multicolumn{5}{l}{\bf Distilled Data}\\
CMLM{~\cite{maskp}}         & 22.54\% & 22.57\% & 2.89\% & 2.70\%\\
\hdashline
~~~+ \textsc{OaXE}~\cite{oaxe}    &  5.01\% & 4.76\% & 1.64\% & 3.00\% \\
GLAT~\cite{glat}                  &  3.16\% & 2.87\% & 1.48\% & 3.32\%\\
~~~+ \textsc{OaXE}~\cite{oaxe}    &  3.06\% & 2.83\% & 1.45\% & 3.15\%\\
DAT~\cite{dat}                    &  0.43\% & 0.27\% & 4.52\% & 5.01\%\\
\bottomrule
\end{tabular}
\caption{Human and automatic evaluation on 100 sentences sampled from WMT20 En$\rightarrow$Zh testset.}
\label{tab:human-automatic}
\end{table*}

\begin{table*}
\centering
\setlength{\tabcolsep}{5pt}
\begin{tabular}{l rrr rrr rr}
\toprule
\multirow{2}{*}{\bf Model}   & \multicolumn{2}{c}{\bf W20 En$\rightarrow$Zh}   & \multicolumn{2}{c}{\bf W20 Zh$\rightarrow$En}   & \multicolumn{2}{c}{\bf W20 En$\rightarrow$De}   & \multicolumn{2}{c}{\bf W20 De$\rightarrow$En}\\
\cmidrule(lr){2-3}\cmidrule(lr){4-5}\cmidrule(lr){6-7}\cmidrule(lr){8-9}
    & \em CRR  & \em DRR    & \em CRR  & \em DRR    & \em CRR  & \em DRR    & \em CRR  & \em DRR\\
\midrule
\multicolumn{7}{l}{\bf Raw Data}\\
CMLM{\small~\cite{maskp}}      & 51.6\% & 1.9\% & 53.5\% & 1.5\% & 40.8\% & 1.6\% & 49.3\% & 1.5\% \\
\hdashline
~~~+ \textsc{OaXE}~\cite{oaxe} &  7.6\% & 4.0\% &  5.9\% & 2.9\% &  3.3\% & 2.8\% &  3.7\% & 3.0\% \\
GLAT~\cite{glat}               &  9.0\% & 3.3\% &  7.0\% & 3.5\% &  3.7\% & 2.2\% &  2.9\% & 2.7\% \\
~~~+ \textsc{OaXE}~\cite{oaxe} &  5.2\% & 3.5\% &  4.4\% & 2.9\% &  2.6\% & 2.6\% &  2.4\% & 2.8\% \\
DAT~\cite{dat}                 &  0.3\% & 7.8\% &  0.1\% & 5.3\% &  0.1\% & 3.5\% &  0.0\% & 3.8\% \\
\midrule
\multicolumn{7}{l}{\bf Distilled Data}\\
CMLM{\small~\cite{maskp}}      & 24.9\% & 2.7\% & 28.6\% & 2.4\% & 17.1\% & 2.3\% & 20.8\% & 2.5\% \\
\hdashline
~~~+ \textsc{OaXE}~\cite{oaxe} &  5.4\% & 3.7\% &  4.9\% & 3.5\% &  2.0\% & 2.6\% &  2.0\% & 2.7\% \\
GLAT~\cite{glat}               &  3.5\% & 3.6\% &  3.1\% & 3.1\% &  1.3\% & 2.4\% &  1.2\% & 2.6\% \\
~~~+ \textsc{OaXE}~\cite{oaxe} &  3.2\% & 3.8\% &  2.7\% & 3.1\% &  1.1\% & 2.4\% &  1.1\% & 2.6\% \\
DAT~\cite{dat}                 &  0.2\% & 5.5\% &  0.1\% & 5.1\% &  0.1\% & 3.1\% &  0.0\% & 3.3\% \\
\bottomrule
\end{tabular}
\caption{Detailed information redundancy benchmarks for fully NAT models on large-scale WMT20 datasets. ``CRR'': continuous redundancy ratio; ``DRR'': , discontinuous redundancy ratio.}
\label{tab:detailed-benchmark}
\end{table*}

\end{document}